\documentclass[runningheads]{llncs}

\usepackage{placeins}
\usepackage{float}
\usepackage{booktabs}
\usepackage{xcolor}
\newif\ifshowtodo
\showtodofalse

\usepackage{amssymb}
\usepackage{amsmath} 
\usepackage{pifont}
\newcommand{\cmark}{\ding{51}}%
\newcommand{\xmark}{\ding{55}}%
\usepackage{hyperref}
\usepackage[all]{hypcap}
\usepackage[T1]{fontenc}
\usepackage{graphicx,verbatim}
\begin{document}
\title{Radiomics--Foundation Fusion for Interpretable RCC Classification: Internal Benchmarking and Exploratory External Transfer}
\titlerunning{Radiomics--Foundation Fusion for RCC Classification}
\author{
Yuan Liang\inst{1,4} \and 
Fangyijie Wang\inst{1,2} \and 
Kathleen M. Curran\inst{1,2} \and 
Gu\'enol\'e Silvestre\inst{1,4} \and
Sourav Bhattacharjee\inst{3} \and 
Abraham Campbell\inst{4}  
}
\authorrunning{Y. Liang et al.}
\institute{
Research Ireland Centre for Research Training in Machine Learning  \and
School of Medicine, University College Dublin, Dublin, Ireland \and
School of Veterinary Medicine, University College Dublin, Dublin, Ireland \and
School of Computer Science, University College Dublin, Dublin, Ireland 
\email{yuan.liang@ucdconnect.ie} \\
}
  
\maketitle             

\begin{abstract}
Accurate preoperative subtype classification of renal cell carcinoma (RCC) from contrast-enhanced CT remains clinically challenging because clear cell RCC (ccRCC) and non-clear cell RCC often show overlapping imaging appearances. This study evaluates whether foundation representations reduce reliance on handcrafted radiomics, or whether radiomics remains complementary for interpretable tumour characterisation. We compared radiomics, conventional CNN features, MedicalNet-pretrained features, MedVAE representations, and fusion variants for binary ccRCC classification on KiTS23, reporting area under the receiver operating characteristic curve (AUC) with bootstrap confidence intervals and average precision (AP) as a complementary class-imbalance-sensitive metric. We further assessed branch-removal ablation, TCGA/AIMI external transfer, and interpretability using radiomics permutation importance and gate-level analysis. Internally, 3D MedVAE gated fusion achieved the best performance, with an AUC of 82.7\% and AP of 92.2\%. On the external TCGA cohort, the same model achieved an AUC of 79.5\% and AP of 98.9\%, although specificity remains uncertain because only two external non-ccRCC cases were available. Gate analysis showed a radiomics-dominant fusion regime, suggesting that foundation representations acted as case-dependent refinement signals rather than replacements for structured tumour descriptors. These findings support radiomics as a complementary and clinically interpretable component of CT-based RCC characterisation in the foundation-model era.


\end{abstract}

\section{Introduction}

Renal cell carcinoma (RCC) is a major urological malignancy for which accurate preoperative subtype characterisation remains clinically important. Distinguishing clear cell RCC (ccRCC) from non-clear cell RCC is clinically relevant because histological subtype is associated with prognosis, treatment response, and management decisions. Contrast-enhanced CT is central to renal mass assessment~\cite{iarc2024kidneyfactsheet,ljungberg2023eau}, but visual interpretation is limited by overlapping enhancement patterns, acquisition heterogeneity, and difficulty in quantifying intratumoural heterogeneity.

Radiomics provides structured tumour descriptors by extracting predefined shape, intensity, and texture features from routine CT~\cite{aerts2014decoding,zwanenburg2020image,van2017computational}. CT radiomics has shown promise for distinguishing ccRCC from non-clear cell subtypes~\cite{wang2021radiomics}, but remains sensitive to segmentation, acquisition protocol, and feature-selection choices. Deep learning and pretrained image representations offer complementary information, with CNN-based transfer learning explored in renal imaging~\cite{litjens2017survey,uhm2021deep,chen2019med3d} and foundation representations increasingly used when labelled data are limited~\cite{bommasani2021opportunities,medvae}. This raises the practical question of whether foundation representations reduce the need for handcrafted radiomics, or whether radiomics remains useful as structured and interpretable tumour evidence.

Interpretability and external transfer are also important for clinical adoption. Post hoc saliency methods can be visually intuitive but weakly coupled to the learned decision pathway~\cite{adebayo2018sanity,ghassemi2021falsehope,vandervelden2022xai,borys2023xai,rudin2019stop}. In contrast, radiomics features correspond to defined tumour phenotype measurements and can support decision-centric interpretation within fusion models. Meanwhile, internal test performance alone is insufficient to assess robustness, motivating cautiously framed external evaluation under acquisition and segmentation shift.

This work investigates radiomics--foundation complementarity for ccRCC versus non-clear cell RCC classification from contrast-enhanced CT. We benchmark radiomics, CNN/MedicalNet features, MedVAE representations, and fusion variants on KiTS23. We further assess TCGA/AIMI external transfer and interpretability using radiomics permutation importance and gate-level analysis.

\section{Methods}

\begin{figure}[t]
\centering
\includegraphics[width=0.9\textwidth]{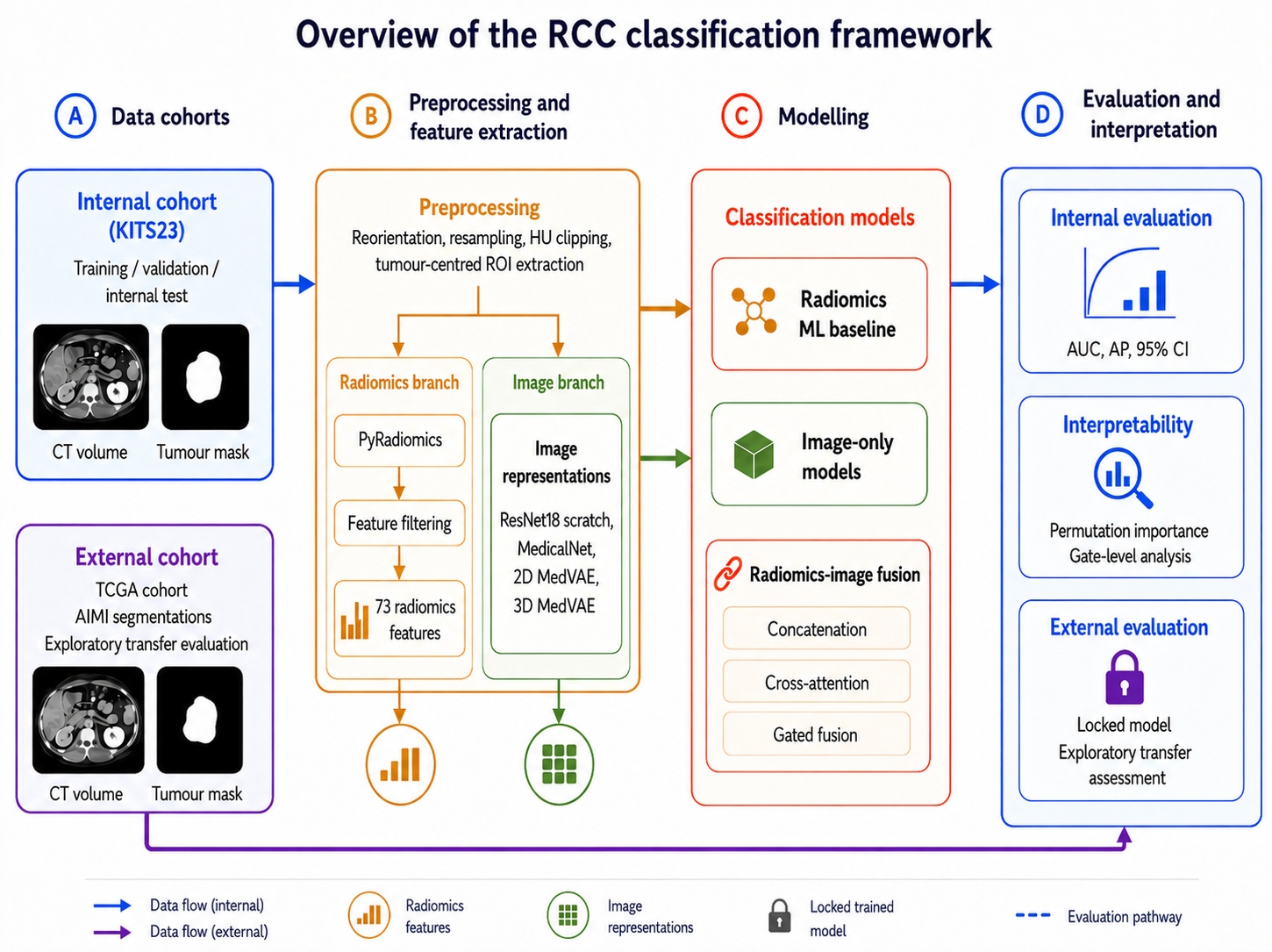}
\caption{Overview of the RCC framework: KiTS23 for internal evaluation, TCGA-KIRC/KIRP/KICH with AIMI segmentations for exploratory external transfer, and radiomics--image modelling with interpretability analyses.}
\label{fig:overview}
\end{figure}

\subsection{Study Cohorts and Preprocessing}

Figure~\ref{fig:overview} summarises the overall study design. The internal cohort was derived from KiTS23, a public multi-institutional contrast-enhanced CT dataset with kidney, tumour, and cyst segmentations~\cite{heller2023kits23}. Multifocal cases were excluded to avoid lesion-level ambiguity, and the final task was formulated as binary ccRCC versus non-clear cell RCC classification.

All CT volumes and masks were reoriented to RAS, resampled to 1.0 mm isotropic spacing, and intensity-clipped to $[-150, 200]$ HU. Image volumes were resampled using linear interpolation and masks using nearest-neighbour interpolation. For each tumour, we generated a zero-margin tumour ROI for radiomics extraction and a tumour-centred ROI with a fixed 12-voxel margin for image-representation learning. When multiple tumour annotations were available, a STAPLE consensus mask was used~\cite{warfield2004staple}.

An independent external cohort was derived from TCGA-KIRC, TCGA-KIRP, and TCGA-KICH. Tumour masks were obtained from the AIMI Annotations initiative, which provides AI-generated DICOM-SEG kidney, tumour, and cyst annotations for NCI Imaging Data Commons collections~\cite{vanoss2023aimi,murugesan2024aigenerated}. To reduce segmentation-quality uncertainty, we retained only expert-corrected masks or masks assigned the highest expert quality score. The external cohort was used only for evaluation, and its positive-enriched composition was interpreted as exploratory transfer evidence rather than definitive diagnostic specificity.

\subsection{Radiomics Feature Extraction and Classical Baseline}

Radiomics features were extracted from the zero-margin tumour ROI using PyRadiomics~\cite{van2017computational} with IBSI-aligned definitions~\cite{zwanenburg2020image}. Shape features were computed from the original image, while first-order and texture features were extracted from original and Laplacian-of-Gaussian filtered images with $\sigma \in \{1,2,3\}$ using a fixed bin width of 25 HU, yielding 386 initial features.

Robustness filtering removed features with ICC $<0.75$, low median absolute deviation, and high pairwise correlation ($|\rho|>0.95$). Correlation-based hierarchical clustering was then used to retain representative features, resulting in a 73-dimensional radiomics representation for fusion models.

For the classical radiomics baseline, the 73-dimensional radiomics representation was defined once from the internal KiTS23 cohort and also used as the radiomics input for fusion models. Classifier-level feature selection, scaling, and SVC fitting were restricted to the internal training data and then applied unchanged to validation, internal test, and external cohorts.

\subsection{Image Representations and Fusion Models}

We compared 3D ResNet-18 trained from scratch, MedicalNet-initialised 3D ResNet-18~\cite{chen2019med3d}, and MedVAE-based image representations~\cite{medvae} under the same preprocessing and evaluation protocol. MedVAE was evaluated using both slice-based 2D and volumetric 3D branches. Image-only models passed the extracted representation directly to a classifier head.

For fusion models, the selected radiomics vector was standardised using training-set statistics and projected to the same latent dimension as the image embedding. We evaluated concatenation, cross-attention, and gated fusion as representative integration strategies. For gated fusion, given image embedding $f_{\mathrm{img}}\in\mathbb{R}^{d}$ and projected radiomics embedding $h_{\mathrm{rad}}\in\mathbb{R}^{d}$, the fused representation was defined as:
\begin{equation}
    g = \sigma(\mathrm{MLP}(h_{\mathrm{rad}})), \quad
    h_{\mathrm{fuse}} = g \odot f_{\mathrm{img}} + (1-g) \odot h_{\mathrm{rad}},
\end{equation}
where $\odot$ denotes element-wise multiplication. Larger gate values indicate greater image-branch weighting, whereas smaller values indicate greater radiomics-branch weighting. This design allowed the model to adaptively modulate image features using structured radiomics evidence, while preserving gate values as an interpretable estimate of branch weighting.

\subsection{Decision-Centric Interpretability}

We performed two complementary interpretability analyses for the best-performing fusion model. First, radiomics permutation importance was used to quantify feature contribution. Each radiomics feature was randomly permuted across cases while keeping image inputs and all other radiomics features unchanged. The procedure was repeated five times per feature, and the mean decrease in AUC was used as the feature-importance score.

Second, we analysed the learned gate values of the gated-fusion model. For each case, the gate vector was extracted during inference and summarised by its mean value. This analysis was performed on both the internal test set and the external cohort to examine how the model weighted image and radiomics branches during prediction.

\section{Experiments and Results}
\subsection{Experimental Setup}
The internal KiTS23 cohort was split at the case level into training, validation, and test sets with fixed proportions of 60\%, 25\%, and 15\%. Model selection used validation AUC. Weighted cross-entropy loss mitigated class imbalance, and image-branch augmentation included random flips, 90-degree rotations, and small affine perturbations; all validation, internal test, and external inputs were processed deterministically.

The primary metric was AUC, with AP, sensitivity, specificity, and other threshold-dependent metrics also reported. Threshold-dependent metrics used a Youden-index threshold selected on the internal validation set and locked for internal test and external TCGA evaluation~\cite{youden1950index}. Bootstrap 95\% confidence intervals were estimated for AUC, and additionally for AP on the external cohort. The TCGA/AIMI cohort was used only for locked-model transfer assessment.

\subsection{Internal Performance on KiTS23}
Table~\ref{tab:internal_performance} summarises the internal held-out test performance. The radiomics SVC baseline achieved an AUC of 74.4\%, outperforming both image-only MedVAE branches. Fusion improved performance over image-only models, with 3D MedVAE concatenation, cross-attention, and gated fusion achieving AUCs of 73.8\%, 79.5\%, and 82.7\%, respectively. The best internal result was obtained by 3D MedVAE gated fusion (AUC 82.7\%, AP 92.2\%). The stronger performance of MedVAE fusion compared with CNN-based fusion may reflect differences in representation structure rather than pretraining alone, with compact generative embeddings complementing radiomics more effectively than discriminative convolutional features in this limited-data setting. The advantage of gated fusion further suggests that case-dependent modality weighting was more effective than fixed concatenation or cross-attention, where less constrained interactions may be more prone to overfitting. Overall, radiomics remained competitive as a standalone descriptor family and provided complementary information when fused with image representations.

\begin{table}[t]
\centering
\caption{Internal held-out test performance for ccRCC versus non-ccRCC classification. AUC includes bootstrap 95\% confidence intervals; all values are percentages.}
\label{tab:internal_performance}
\resizebox{\textwidth}{!}{
\begin{tabular}{llcccccc}
\toprule
Category & Model & AUC (95\% CI) & AP & Accuracy & F1 & Sensitivity & Specificity \\
\midrule
Radiomics & SVC baseline & 74.4 [61.2, 87.3] & 84.8 & 62.0 & 65.6 & 52.6 & 83.8 \\
CNN fusion & ResNet18 gated fusion & 78.4 [63.5, 90.9] & 86.5 & 75.0 & 81.9 & 81.0 & 61.1 \\
CNN fusion & MedicalNet gated fusion & 78.6 [63.9, 91.3] & 86.4 & 70.0 & 75.0 & 64.3 & 83.3 \\
Foundation only & 2D MedVAE & 59.5 [42.9, 75.4] & 78.8 & 63.3 & 75.0 & 78.6 & 27.8 \\
Foundation only & 3D MedVAE & 63.2 [46.0, 79.9] & 77.2 & 71.7 & 82.5 & 95.2 & 16.7 \\
Foundation fusion & 2D MedVAE gated fusion & 79.6 [67.0, 90.4] & 90.1 & 76.7 & 84.4 & 90.5 & 44.4 \\
Foundation fusion & 3D MedVAE concatenation & 73.8 [60.5, 87.3] & 86.3 & 61.7 & 67.6 & 57.1 & 72.2 \\
Foundation fusion & 3D MedVAE cross-attention & 79.5 [67.6, 90.5] & 88.9 & 71.7 & 79.5 & 78.6 & 55.6 \\
Foundation fusion & 3D MedVAE gated fusion & \textbf{82.7 [70.7, 92.2]} & \textbf{92.2} & 71.7 & 77.9 & 71.4 & 72.2 \\
\bottomrule
\end{tabular}
}
\end{table}

\subsection{Branch-Removal Ablation of the Best Fusion Model}
We further performed branch-removal ablation on the best-performing 3D MedVAE gated-fusion model by routing either the image representation or the projected radiomics representation alone through the trained gated-fusion classifier, while keeping all model parameters fixed. As shown in Table~\ref{tab:ablation}, the full fusion model achieved an AUC of 82.7\%, whereas image-branch-only and radiomics-branch-only inference reached 60.4\% and 52.9\%, respectively. This supports radiomics--image complementarity within the trained fusion pathway. These single-branch variants are not independently optimised classifiers, and therefore assess pathway dependence rather than standalone modality performance. The large gap between full fusion and either branch alone indicates that the classifier did not simply rely on a single dominant input, but required joint access to both representation spaces.

\begin{table}[ht]
\centering
\caption{Branch-removal ablation within the trained 3D MedVAE gated-fusion model. AUC is reported on the internal test set.}
\label{tab:ablation}
\scalebox{0.8}{
\setlength{\tabcolsep}{5pt}
\begin{tabular}{lccc}
\toprule
Variant & Image branch & Radiomics branch & AUC (\%) \\
\midrule
Radiomics branch only & \xmark & \cmark & 52.9 \\
Image branch only & \cmark & \xmark & 60.4 \\
Full fusion & \cmark & \cmark & \textbf{82.7} \\
\bottomrule
\end{tabular}
}
\end{table}
\subsection{Exploratory External Transfer on TCGA with AIMI Segmentations}

Table~\ref{tab:external_performance} reports locked-model external evaluation on the TCGA renal cancer cohort with AIMI segmentations. This analysis was intended as exploratory transfer assessment rather than definitive external validation, because the cohort was strongly positive-enriched and contained only two non-ccRCC cases. Therefore, specificity and AUC should be interpreted cautiously, and confusion-matrix counts are reported explicitly.

\begin{table}[ht]
\centering
\caption{Exploratory external transfer performance on the TCGA cohort with AIMI segmentations. Models were evaluated using locked internal-validation thresholds without external refitting. AUC and AP are reported with bootstrap 95\% confidence intervals. All metric values are percentages.}
\label{tab:external_performance}
\resizebox{\textwidth}{!}{
\begin{tabular}{llccccccc}
\toprule
Category & Model & $n$ & AUC (95\% CI) & AP (95\% CI) & F1 & Sensitivity & Specificity & TN/FP/FN/TP \\
\midrule
Radiomics & SVC baseline & 41 & 83.3 [59.0, 100.0] & 99.0 [97.5, 100.0] & 97.4 & 97.4 & 50.0 & 1/1/1/38 \\
CNN fusion & ResNet18 gated fusion & 41 & 93.6 [79.5, 100.0] & 99.7 [98.9, 100.0] & 93.3 & 89.7 & 50.0 & 1/1/4/35 \\
CNN fusion & MedicalNet gated fusion & 41 & 96.2 [87.2, 100.0] & 99.8 [99.3, 100.0] & 90.1 & 82.1 & 100.0 & 2/0/7/32 \\
Foundation only & 2D MedVAE & 41 & 70.5 [30.8, 100.0] & 97.9 [94.7, 100.0] & 87.3 & 79.5 & 50.0 & 1/1/8/31 \\
Foundation only & 3D MedVAE & 41 & 64.1 [33.3, 92.3] & 97.6 [95.1, 99.6] & 92.1 & 89.7 & 0.0 & 0/2/4/35 \\
Foundation fusion & 2D MedVAE gated fusion & 41 & 83.3 [65.4, 97.4] & 99.1 [98.0, 99.9] & 93.3 & 89.7 & 50.0 & 1/1/4/35 \\
Foundation fusion & 3D MedVAE gated fusion & 41 & 79.5 [66.6, 92.3] & 98.9 [98.1, 99.6] & 87.0 & 76.9 & 100.0 & 2/0/9/30 \\
\bottomrule
\end{tabular}
}
\end{table}

Several internally trained models retained high positive-class ranking under TCGA/AIMI domain and segmentation shift. The 3D MedVAE gated-fusion model achieved an external AUC of 79.5\% and AP of 98.9\%, while MedicalNet gated fusion achieved the highest external AUC of 96.2\%. This external ranking should be interpreted cautiously, as the apparent MedicalNet advantage may reflect model robustness under TCGA/AIMI shift but was estimated from only two negative cases. This may indicate that different pretrained representations respond differently to acquisition and segmentation shift, rather than a stable superiority of one model family. High AP values mainly reflect positive-class ranking in a cohort with 95.1\% ccRCC prevalence, and the very small number of negative cases limits conclusions about balanced diagnostic performance. These results therefore support exploratory transfer feasibility, but not definitive population-level generalisation.

\subsection{Radiomics-Based Interpretability}

\begin{figure}[ht]
    \centering
    \includegraphics[width=0.827\textwidth]{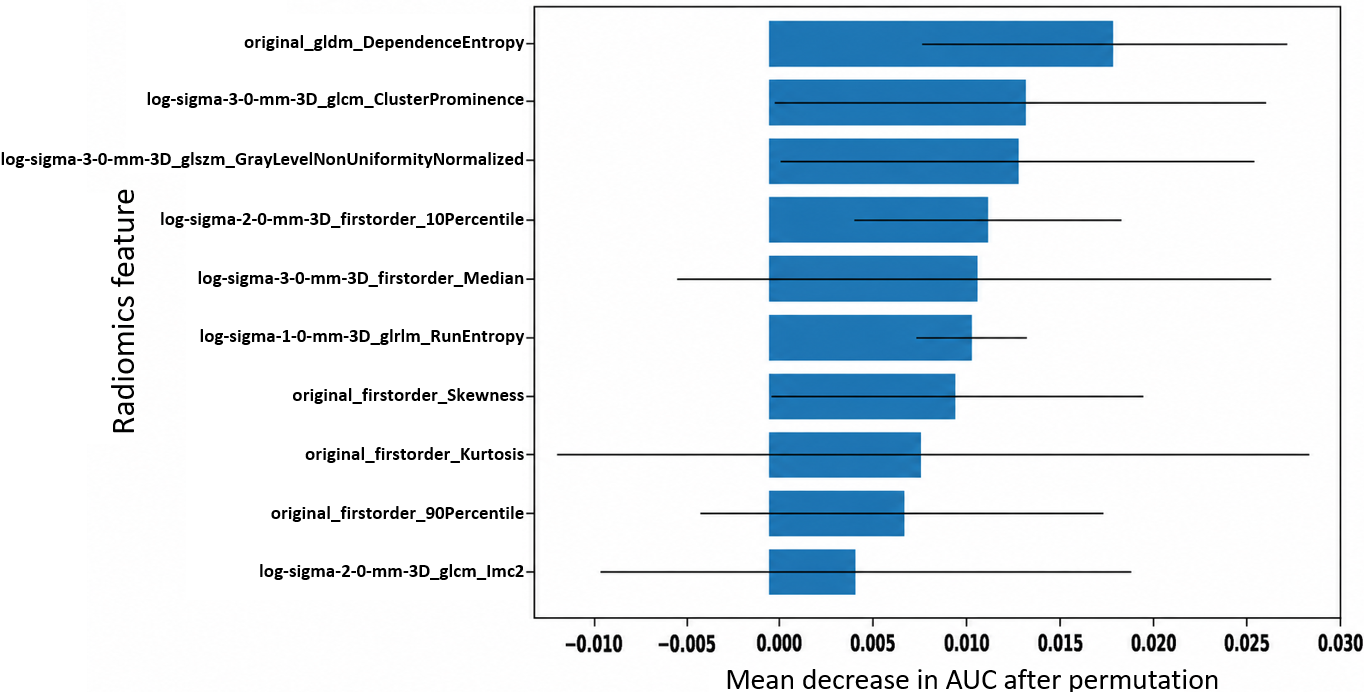}
    \caption{Top radiomics features ranked by mean AUC decrease after repeated feature-wise permutation. Error bars show variation across repetitions.}
    \label{fig:perm_importance}
\end{figure}

To examine which structured tumour descriptors contributed most strongly to the best-performing fusion model, we performed repeated permutation importance analysis on the radiomics branch. Figure~\ref{fig:perm_importance} shows the top-ranked radiomics features according to the mean decrease in AUC after feature-wise permutation. The most influential descriptors were dominated by texture and heterogeneity-related feature families, including GLDM, GLCM, GLSZM, GLRLM, first-order, and Laplacian-of-Gaussian-derived features.

These results indicate that the fusion model relied on interpretable descriptors of intratumoural heterogeneity and intensity distribution, rather than only on abstract image embeddings. Because these features are defined radiomics measurements, they provide an imaging-level interpretation of model behaviour. However, they should be interpreted as tumour phenotype descriptors rather than direct histopathological surrogates.

\subsection{Gate-Level Interpretability}

We analysed gate values from the 3D MedVAE gated-fusion model. Each case produced a 512-dimensional gate vector, where larger values indicate greater image-branch weighting and smaller values indicate greater radiomics-branch weighting. The analysis was performed on the internal test set and on the 41-case external TCGA cohort with AIMI segmentations, which contained 39 ccRCC and 2 non-ccRCC cases.

\begin{figure}[ht]
    \centering
    \includegraphics[width=0.78\textwidth]{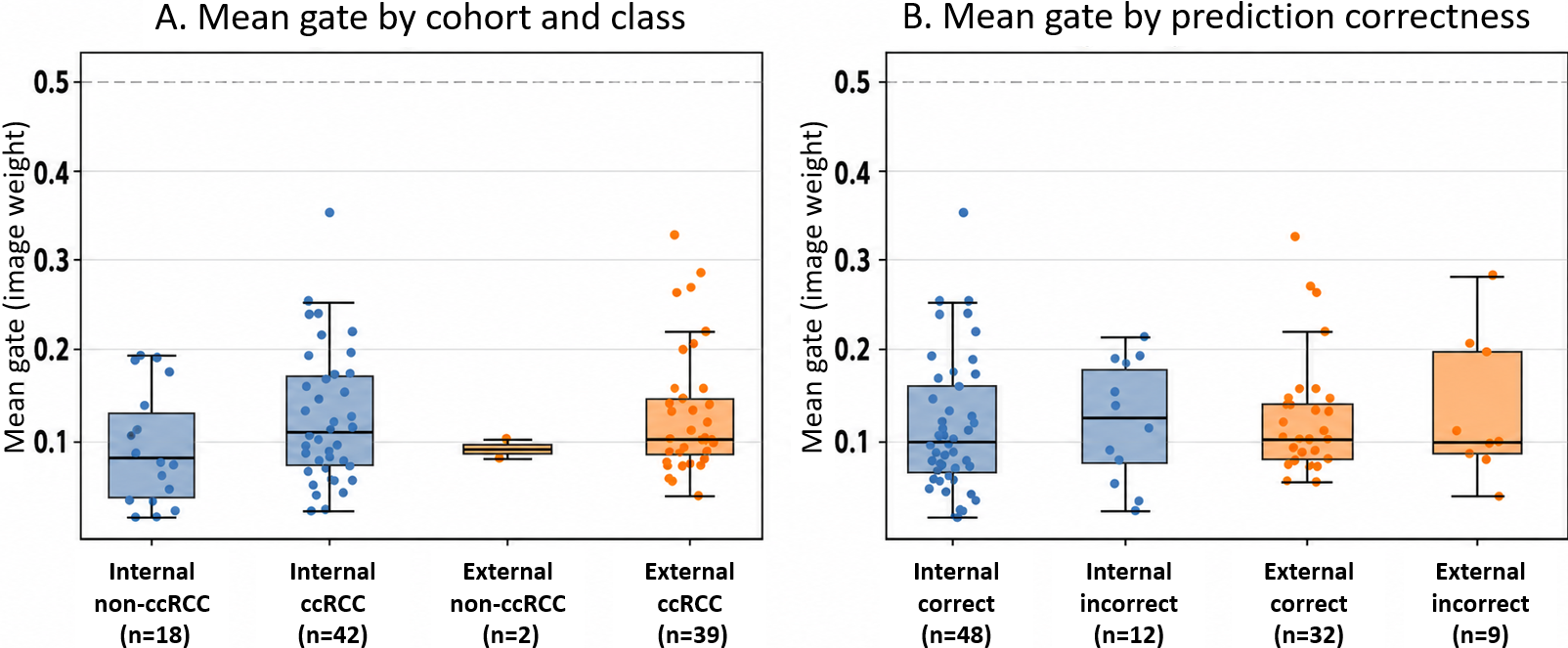}
    \caption{Gate-level analysis of the 3D MedVAE gated-fusion model. Each point shows the case-level mean of the 512-dimensional gate vector; higher values indicate greater image weighting and lower values greater radiomics weighting. External class-stratified patterns should be interpreted cautiously due to only two non-ccRCC cases.}

    \label{fig:gate_analysis}
\end{figure}

As shown in Fig.~\ref{fig:gate_analysis}, mean gate values were consistently below 0.5 in both cohorts. The internal test set showed a mean gate value of 0.1227 $\pm$ 0.0675, while the external cohort showed a similar mean value of 0.1328 $\pm$ 0.0619. This indicates that the best-performing fusion model operated in a radiomics-dominant regime rather than using balanced multimodal weighting.

Because the external cohort contained only two non-ccRCC cases, class-stratified gate comparisons on the external cohort should be interpreted cautiously. Nevertheless, the overall gate distribution suggests that the model primarily relied on structured radiomics descriptors, with image representations acting as case-dependent modulation signals.

\section{Conclusion}

This study shows that handcrafted radiomics remains clinically relevant for CT-based RCC subtype classification in the foundation-model era. Radiomics was competitive as a standalone descriptor family, while branch-removal ablation and internal performance indicated that the best 3D MedVAE fusion pathway depended on both radiomics and image representations, supporting complementary rather than substitutive value. Exploratory TCGA/AIMI evaluation suggested transfer feasibility under acquisition and segmentation shift, but the positive-enriched external cohort and limited non-ccRCC cases preclude definitive claims about diagnostic specificity. Gate-level and permutation analyses further showed a radiomics-dominant decision pathway, supporting structured tumour descriptors as clinically communicable model evidence. Larger balanced external cohorts and prospective validation remain necessary before clinical deployment.

\begin{credits}
\subsubsection{\ackname} Yuan Liang acknowledges doctoral scholarship support from the China Scholarship Council (CSC). This work was funded by Taighde \'{E}ireann -- Research Ireland through the Research Ireland Centre for Research Training in Machine Learning (18/CRT/6183).
\subsubsection{\discintname}
The authors have no competing interests to declare that are relevant to the content of this article. 
\end{credits}

\newpage

\bibliographystyle{splncs04}
\bibliography{bib}

\end{document}